%% file: main.tex
\documentclass{article}
\usepackage{iclr2027_conference,times}

\input{math_commands.tex}

\usepackage{amsmath,amssymb}
\usepackage{graphicx}
\usepackage{algorithm}
\usepackage{algpseudocode}
\usepackage{booktabs}
\usepackage{caption}
\usepackage{comment}
\usepackage{multirow}
\usepackage{hyperref}
\usepackage{url}
\hypersetup{hidelinks}
\setcitestyle{numbers,square}

\title{Ember: A Kilobyte-State Optimizer for Token Interfaces}
\author{Kathan Shah}
\iclrfinalcopy

\begin{document}
\raggedbottom
\maketitle
\lhead{Preprint}
\input{content_iclr}
\end{document}

%% file: math_commands.tex
\usepackage{amsmath,amsfonts,bm}

\def\eqref#1{equation~\ref{#1}}

\def\1{\bm{1}}

\DeclareMathAlphabet{\mathsfit}{\encodingdefault}{\sfdefault}{m}{sl}
\SetMathAlphabet{\mathsfit}{bold}{\encodingdefault}{\sfdefault}{bx}{n}



%% file: content_iclr.tex
\begin{abstract}
Language models learn continuous programs over discrete symbols, with the embedding table and LM-head acting as the read/write interface between them.
We show that this interface has gradient geometry distinct from dense hidden weights which can be exploited to improve the Pareto frontier across supervised finetuning, RL, and pretraining, while only utilizing kilobytes of optimizer state.
We introduce \emph{Ember}, a lightweight optimizer for embedding and LM-head matrices that utilizes $\mathcal{O}(V + D)$ optimizer VRAM, instead of Adam's $\mathcal{O}(2VD)$, and forgoes the need to shard both token table optimizer states. 
On 50k-step pretraining across two datasets and three paired seeds, Ember lowers val loss by \(0.08\)–\(0.09\) nats relative to tuned AdamW; we also provide empirical evidence that Ember scales across batch size and parameter count.
Finally, we provide a distributed Ember implementation that merges cleanly with existing ZeRO/FSDP setups.

\end{abstract}

\begin{center}
\includegraphics[width=\linewidth]{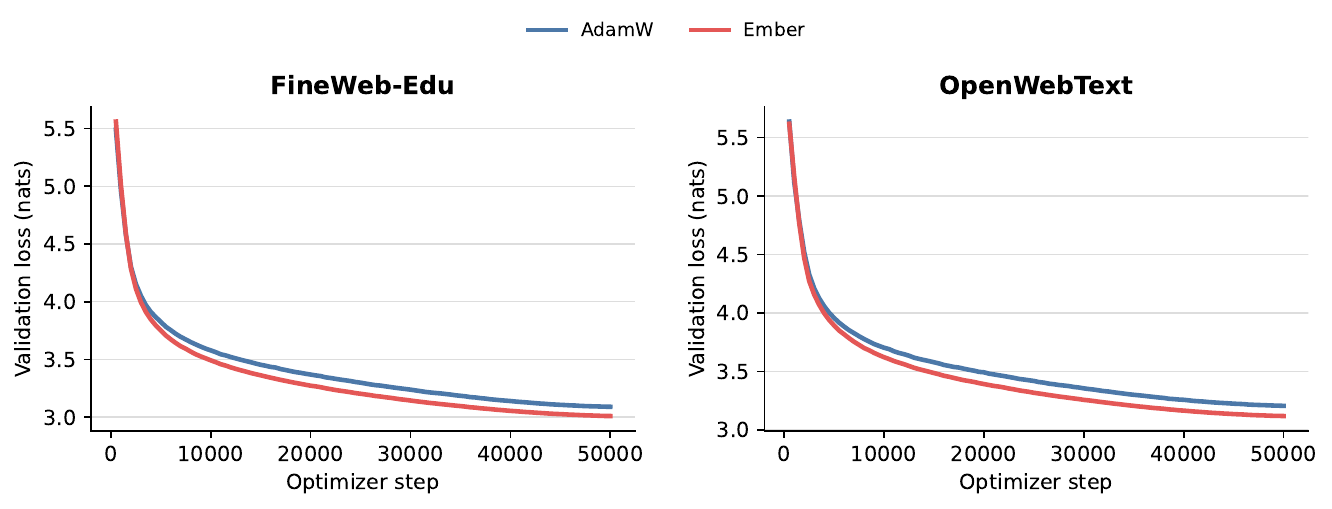}
  \captionof{figure}{\textbf{Long-horizon pretraining.} Validation loss on FineWeb-Edu and OpenWebText over 50,000 optimizer steps; curves are means over three paired seeds. Only the optimizer on the untied input embedding and LM head changes, while the Transformer body uses the same AdamW setup in every run.}
\label{fig:headline_trajectory}
\end{center}

\section{Introduction}

Embedding and LM-head matrices, or ``token interface'' parameters, are currently optimized with Adam \cite{kingma2014adam}. This is corroborated by a report from the current open source frontier, DeepSeek V4 \cite{deepseek2026v4}, which only changed the optimizer for linear layers to Muon \cite{jordan2024muon}. 
Of note, almost all recent proposed successors of Adam solely apply to linear layers and steer clear of the embedding table. 
Therefore improvements are pertinent as Adam requires significant distributed training engineering and doesn't fully leverage the geometry of token interfaces. In this paper, we introduce \emph{Ember}, a lightweight optimizer for such Transformer parameters that remedies these and achieves competitive results.

\section{Methods}

\subsection{Derivation}

We drop Adam's first moment and approximate its second-moment buffer with an outer product of row and column factors (Alg.~\ref{alg:ember}, Fig.~\ref{fig:outerproduct_residual}). Beyond the learning rate, the only hyperparameter is $\beta_2=0.999$, which we lift directly from the original Adam paper. The all-row column-averaging variant is equivalent to Algorithm~2 in \cite{shazeer2018adafactor} under matched settings and away from numerical guards. The headline 50k-step experiments instead average columns over active rows, as specified in Algorithm~\ref{alg:ember}; this generally changes the update for sparse gradients. Our contribution is its evaluation and implementation for token interfaces while retaining the body optimizer. We discuss the relationship in Related Work and compare with native Adafactor in Appendix~\ref{sec:adafactor_comparison}.

We begin from Adam's second-moment buffer: an exponential moving average of squared minibatch gradients. Fisher geometry motivates this form, but this statistic is not generally the model Fisher or the average of per-example gradient squares.

Linear-layer gradients from backpropagation have an outer-product structure. For a single-position forward pass $y=xW$, the weight gradient is $dW=x^\top dy$, an outer product of rank at most one. In an input embedding, $x$ is one-hot, so each position contributes to just one token row. Summing over a batch preserves this row assignment, although the summed gradient need not remain rank one.

Intuitively, backpropagation accumulates gradients every time a token is selected. A token with higher frequency can therefore have a gradient $\ell_2$ norm roughly proportional to its frequency, assuming its contributions have comparable magnitudes and point in similar directions. Measured row-gradient energy is strongly associated with token frequency ($r=0.976$). Ember measures each row's squared gradients directly rather than assuming a frequency law; Appendix~\ref{app:inverse_probability_curvature} gives the mean and variance terms.

Row-only scaling assigns the same second-moment estimate to every feature in a token row. We add an $\mathcal{O}(D)$ column factor to capture feature-wise differences (Fig.~\ref{fig:outerproduct_residual}, Section~\ref{app:decomposing_squared_gradient}). The default algorithm and headline experiments use both factors; we use the row-only version to explain token frequency.

The product of the two factors has units of squared gradient squared. Geometric-mean normalization restores the units of a second moment, apart from the numerical floor (App.~\ref{app:outer_product_unit_alignment}). It equals a common grand mean only for the all-row variant, not generally for active-row column averaging.

Removing the first-moment EMA saves another $V \times D$ state values per token table. Adding it back has no consistent benefit in our two tested batch settings (Table~\ref{tab:decomp}); Figure~\ref{fig:no_first_mom} reports the gradient-autocorrelation diagnostic.

\begin{algorithm}[t]
\caption{\textit{Ember} for embedding and LM-head matrices: the active-row column-averaging implementation used in the headline 50k-step experiments. We use $\beta_2 = 0.999$; learning rates are specified for each experiment.}
\label{alg:ember}
\begin{algorithmic}[1]
\Require $\alpha$: step size
\Require $\beta_2 \in [0,1)$: momentum term for second-moment estimates
\Require $R_t(\theta)$: reward at timestep $t$ (WLOG $R_t = -\mathcal{L}_t$ for loss minimization)
\Require $\theta_0 \in \mathbb{R}^{V \times D}$: initial parameters of the embedding or LM-head matrix

\State $r_0 \leftarrow \mathbf{0} \in \mathbb{R}^{V}$
\Comment{Initialize row second-moment vector}

\State $c_0 \leftarrow \mathbf{0} \in \mathbb{R}^{D}$
\Comment{Initialize column second-moment vector}

\State $t \leftarrow 0$
\Comment{Initialize timestep}

\While{$\theta_t$ not converged}
    \State $t \leftarrow t + 1$

    \State $g_t \leftarrow \nabla_{\theta} R_t(\theta_{t-1})$
    \Comment{Compute gradient through backprop}

    \State $a_t \leftarrow \max(1, |\{i:\exists j,\ g_{t,ij}\ne 0\}|)$
    \Comment{Count active gradient rows}

    \State $r_t \leftarrow
    \beta_2 r_{t-1}
    + (1-\beta_2)\operatorname{mean}_{j}(g_t^2)$
    \Comment{Update row second moment}

    \State $c_t \leftarrow
    \beta_2 c_{t-1}
    + (1-\beta_2)
    a_t^{-1}\sum_i g_{t,i:}^2$
    \Comment{Average over active rows}

    \State $\hat r_t \leftarrow r_t/(1-\beta_2^t)$
    \Comment{Bias-correct row estimate}

    \State $\hat c_t \leftarrow c_t/(1-\beta_2^t)$
    \Comment{Bias-correct column estimate}

    \State $s_t \leftarrow
    \sqrt{\max(\operatorname{mean}(\hat r_t)\,\operatorname{mean}(\hat c_t),10^{-30})}$
    \Comment{Geometric-mean normalization}

    \State $\tilde v_t \leftarrow
    \hat r_t \hat c_t^\top / s_t$
    \Comment{Form factored preconditioner}

    \State
    $\theta_t \leftarrow
    \theta_{t-1}
    + \alpha
    \, g_t /
    (\sqrt{\tilde v_t} + 10^{-8})$
    \Comment{Ascend the reward $R_t$}

\EndWhile

\State \Return $\theta_t$
\end{algorithmic}
\end{algorithm}

\subsection{Optimizing sparsely activated parameters}

When a row's gradient and RMS both scale with token frequency $p$, unnormalized updates scale with $p$, inverse-second-moment updates with $1/p$, and inverse-RMS updates are approximately frequency-independent. Ember's row factor uses the last normalization; Appendix~\ref{app:scaling_views} explains this approximation.

\section{Results}

\subsection{Pretraining}

\textbf{Long-horizon pretraining.}
We trained Pythia-410M from scratch for 50,000 steps on FineWeb-Edu and OpenWebText (1.6384 billion tokens per run). For each corpus, AdamW and Ember's token-interface learning rates were selected independently from the same four-point grid before evaluation on three fresh paired seeds per method. At the frozen endpoint --- the mean of 11 validation measurements from steps 45{,}000--50{,}000 --- Ember improved held-out cross-entropy by $0.0802$ nats on FineWeb-Edu and $0.0882$ nats on OpenWebText, and was lower on every paired seed (Fig.~\ref{fig:headline_trajectory}). At the shared token-table shape, Ember removes approximately $824$ MB of FP32 token-interface optimizer state. Full protocol and paired confidence intervals are in Appendix~\ref{app:long_horizon_410m}.

\begin{figure}[H]
\centering
\includegraphics[width=\linewidth]{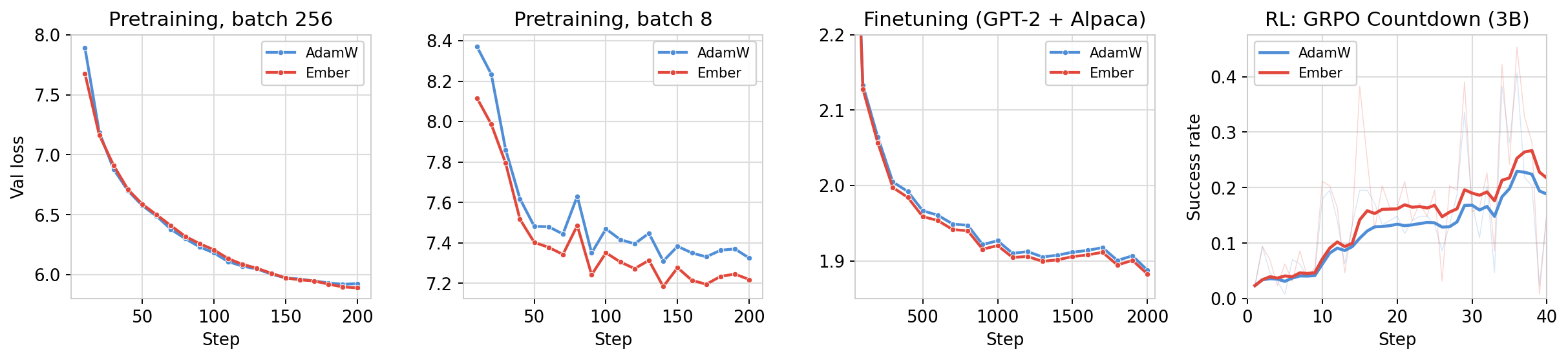}
\caption{Overview of earlier experiments. Panels A and B compare pretraining at batch sizes 256 and 8, panel C shows supervised fine-tuning, and panel D shows GRPO post-training. Curves show representative runs; replicated endpoint comparisons are reported in the text and tables.}
\label{fig:breadth_overview}
\end{figure}

\textbf{Small-scale pretraining.}
In a separate 200-step GPT-2-small experiment with Muon on the linear layers and $n{=}3$ paired seeds, Ember matches or outperforms AdamW and improves over canonical Adafactor by $0.095$ nats at batch 8 and $0.251$ nats at batch 256, while retaining the same $\mathcal{O}(V{+}D)$ state complexity (Fig.~\ref{fig:breadth_overview}, Table~\ref{tab:decomp}).

\subsection{Model-size robustness}

\textbf{Model size.}
We compare model sizes using 250-step supervised fine-tuning, rather than long-horizon pretraining.
Ember matches AdamW as model size increases while removing the $O(2VD)$ optimizer state on the 2 token tables (Fig.~\ref{fig:model_scaling}). On GPT-2, Ember improves validation loss at every scale. On Pythia, the mean loss difference favors Ember at 160M and is near zero at 1.4B and 2.8B (Table~\ref{tab:model_scaling_combined}).

\begin{figure}[!htbp]
\centering
\includegraphics[width=0.66\linewidth]{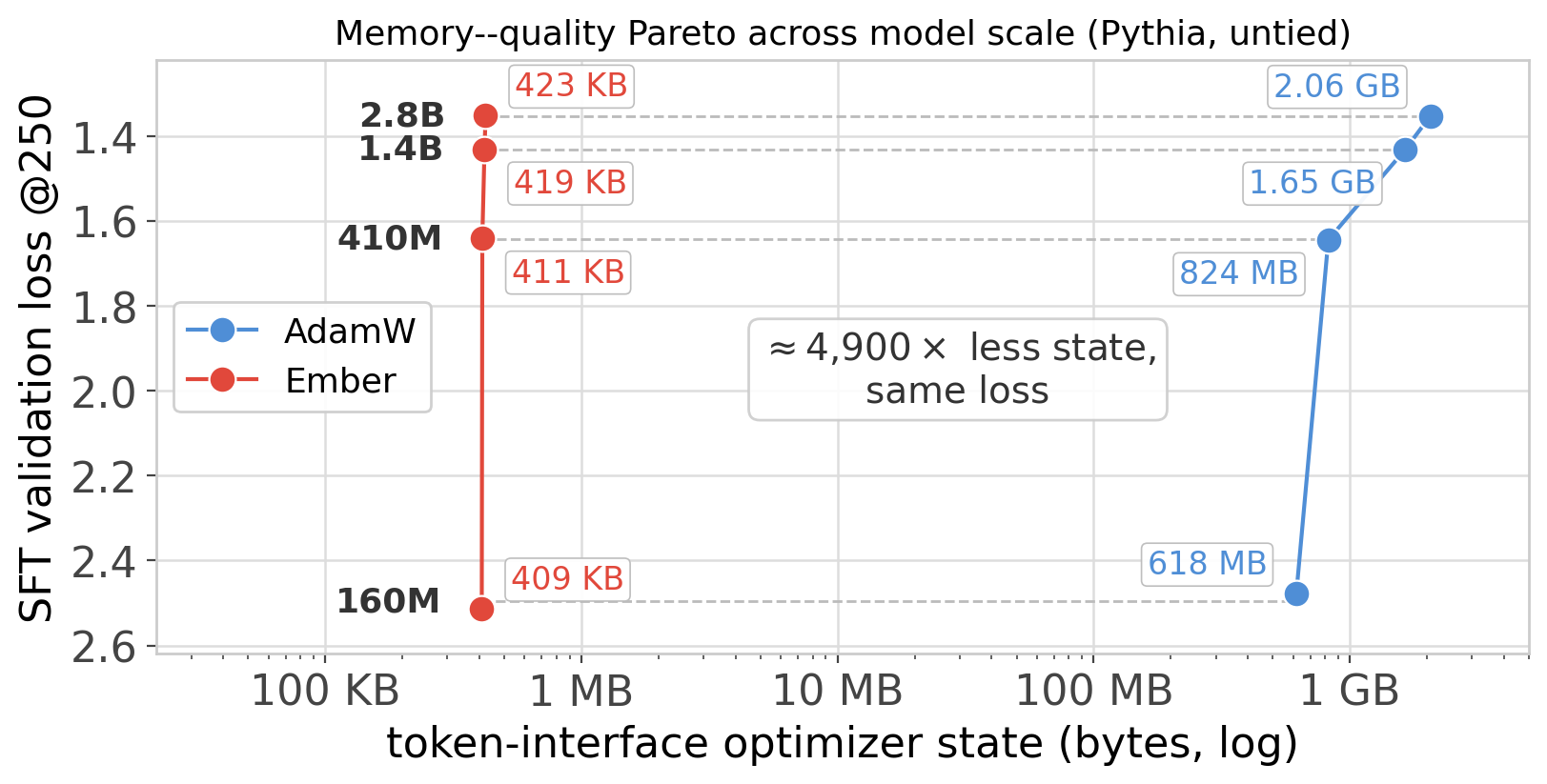}
\caption{\textbf{The memory--quality gap widens with model scale.} The two optimizers reach the same loss at every scale, yet AdamW's state grows as $\mathcal{O}(2VD)$ --- crossing $1$\,GB by 1.4B and reaching
$\sim\!2$\,GB at 2.8B --- while Ember's $\mathcal{O}(V{+}D)$ state stays near $400$\,KB.}
\label{fig:memory_pareto_scale}
\end{figure}

We visually depict the optimizer state benefit in Fig.~\ref{fig:memory_pareto_scale}.
Notably, on Pythia-2.8B, AdamW's token-interface state reaches $2$\,GB, whereas Ember
stays at $400$\,KB. This is a $4{,}900\times$ reduction for the same val loss.

\begin{table}[t]
\centering
\caption{Supervised fine-tuning validation loss at step 250 across GPT-2 and Pythia model scales,
with per-model embedding-optimizer state. Ember is slightly ahead on GPT-2 and within seed noise on Pythia, while reducing token-interface optimizer state from $\mathcal{O}(2VD)$ to
$\mathcal{O}(V{+}D)$. State is FP32 optimizer buffers.}
\label{tab:model_scaling_combined}
\small\setlength{\tabcolsep}{3pt}
\begin{tabular}{lrcccrr}
\toprule
Model & Params (M) & AdamW & Ember & $\Delta$ & AdamW state & Ember state \\
\midrule
distilgpt2  &   82 & 2.3670{\scriptsize$\pm$.0076} & \textbf{2.3611{\scriptsize$\pm$.0072}} & $\mathbf{-0.0059{\scriptstyle\pm}.0004}$ & 309 MB  & 199 KB \\
gpt2        &  124 & 2.0529{\scriptsize$\pm$.0030} & \textbf{2.0450{\scriptsize$\pm$.0032}} & $\mathbf{-0.0079{\scriptstyle\pm}.0002}$ & 309 MB  & 199 KB \\
gpt2-medium &  350 & 1.7483{\scriptsize$\pm$.0033} & \textbf{1.7452{\scriptsize$\pm$.0033}} & $\mathbf{-0.0032{\scriptstyle\pm}.0001}$ & 412 MB  & 200 KB \\
gpt2-large  &  774 & 1.5937{\scriptsize$\pm$.0074} & \textbf{1.5925{\scriptsize$\pm$.0075}} & $\mathbf{-0.0012{\scriptstyle\pm}.0001}$ & 515 MB  & 201 KB \\
\midrule
pythia-160m &  160 & 2.5306{\scriptsize$\pm$.0284} & 2.5175{\scriptsize$\pm$.0325} & $-0.0132{\scriptstyle\pm}.0318$ & 618 MB  & 399 KB \\
pythia-1.4b & 1400 & 1.4298{\scriptsize$\pm$.0040} & 1.4298{\scriptsize$\pm$.0039} & $-0.0001{\scriptstyle\pm}.0003$ & 1.65 GB & 409 KB \\
pythia-2.8b & 2800 & 1.3530{\scriptsize$\pm$.0025} & 1.3525{\scriptsize$\pm$.0026} & $-0.0005{\scriptstyle\pm}.0006$ & 2.06 GB & 413 KB \\
\bottomrule
\multicolumn{7}{l}{\scriptsize Mean $\pm$ sd, $n{=}3$ paired seeds; bold: $|\Delta| > 2\sigma$.}
\end{tabular}
\end{table}

\begin{figure}[H]
  \centering
  \includegraphics[width=\linewidth]{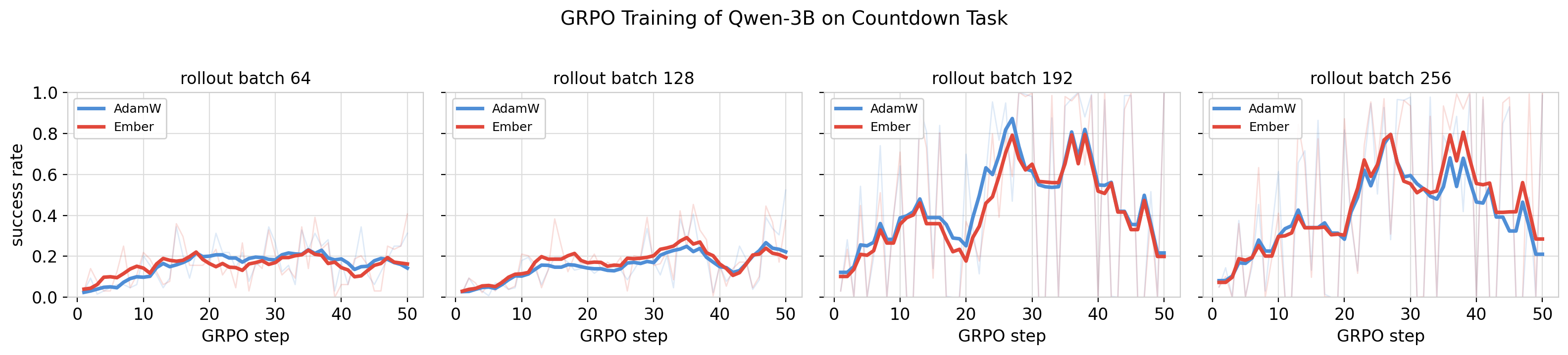}
  \caption{\textbf{Ember is competitive in reinforcement learning while removing
  $\mathcal{O}(2VD)$ optimizer state.} Comparisons at rollout batch sizes $64$, $128$, $192$, and $256$. Batches 64 and 256 are replicated over $n{=}3$ seeds; the intermediate sizes show representative runs.}
  \label{fig:rl_countdown}
\end{figure}

\textbf{Reinforcement learning.} We compare Ember against AdamW on the token tables when post-training
Qwen2.5-3B-Instruct with GRPO~\cite{shao2024deepseekmath} on the Countdown
reasoning task, sweeping the rollout batch size from $64$ to $256$
(Fig.~\ref{fig:rl_countdown}). The curves are comparable across the tested batch sizes; the endpoints at batches 64 and 256 are replicated over three seeds. Ember uses only $600$ KB of optimizer
state compared to AdamW's $2.5$ GB (a $4200\times$ reduction).

\subsection{Ablations}
\begin{figure}[H]
\centering
\includegraphics[width=\linewidth]{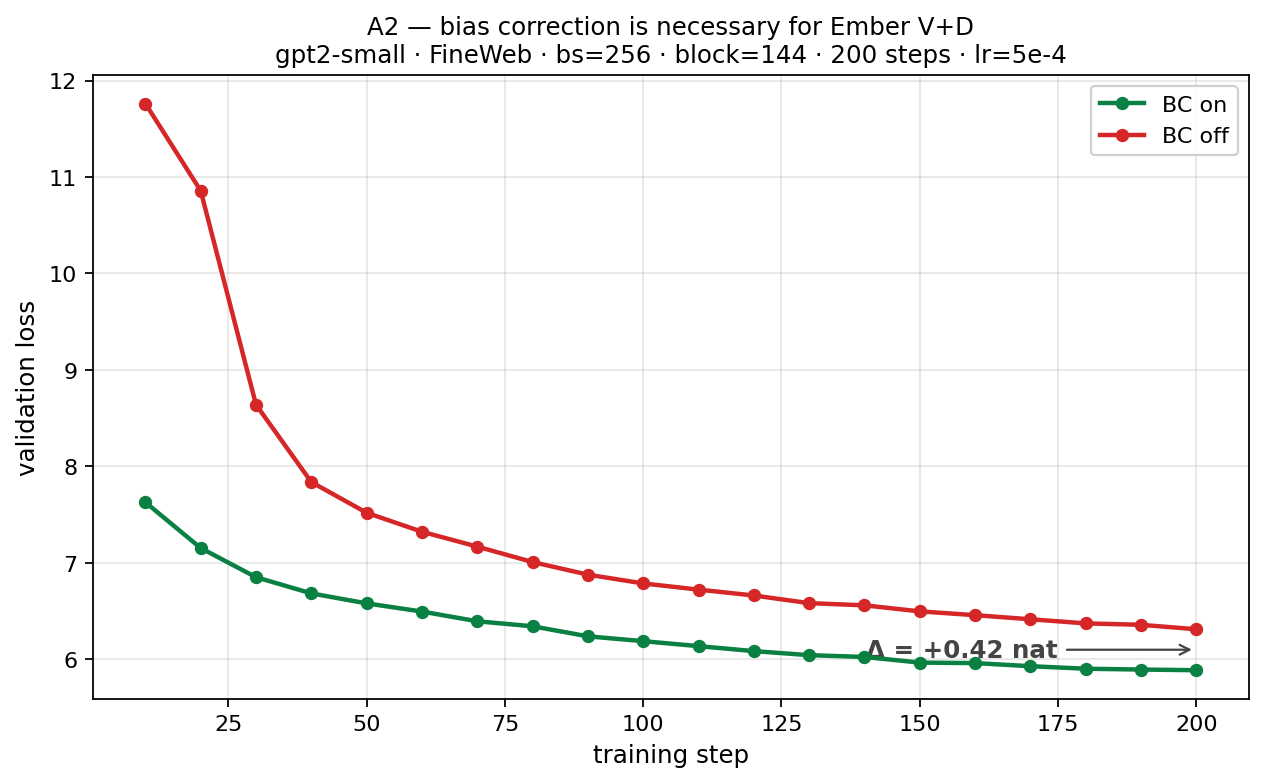}
\caption{\textbf{Ablating bias correction in Ember.} Including the bias correction is crucial to performance and makes training much stabler while helping the optimizer reach lower val loss faster.}
\label{fig:bc_failure}
\end{figure}

\textbf{Bias correction.}
Bias correction improves stability and validation loss in the tested configuration (Fig.~\ref{fig:bc_failure}).

\subsection{Decomposing the Squared Gradient}
\label{app:decomposing_squared_gradient}

For a single token position, write the gradient contribution as
\[
Z_{ij}=\delta_i x_j,\qquad Z=\delta x^\top.
\]
For an input embedding, $\delta$ denotes the one-hot row selector and $x$ the backpropagated feature gradient. For the LM head, $\delta$ is the softmax error and $x$ the hidden state.

If the squared factors are independent, the per-position second moment separates:
\[
\mathbb{E}[Z_{ij}^2]
=\mathbb{E}[\delta_i^2]\mathbb{E}[x_j^2].
\]
Without independence, the difference is $\operatorname{Cov}(\delta_i^2,x_j^2)$. This motivates a factored approximation, but does not prove it for the minibatch statistic. A batch gradient $G=\frac{1}{M}\sum_a Z_a$ contains cross-position terms when squared.

\begin{figure}[t]
\centering
\includegraphics[width=\linewidth]{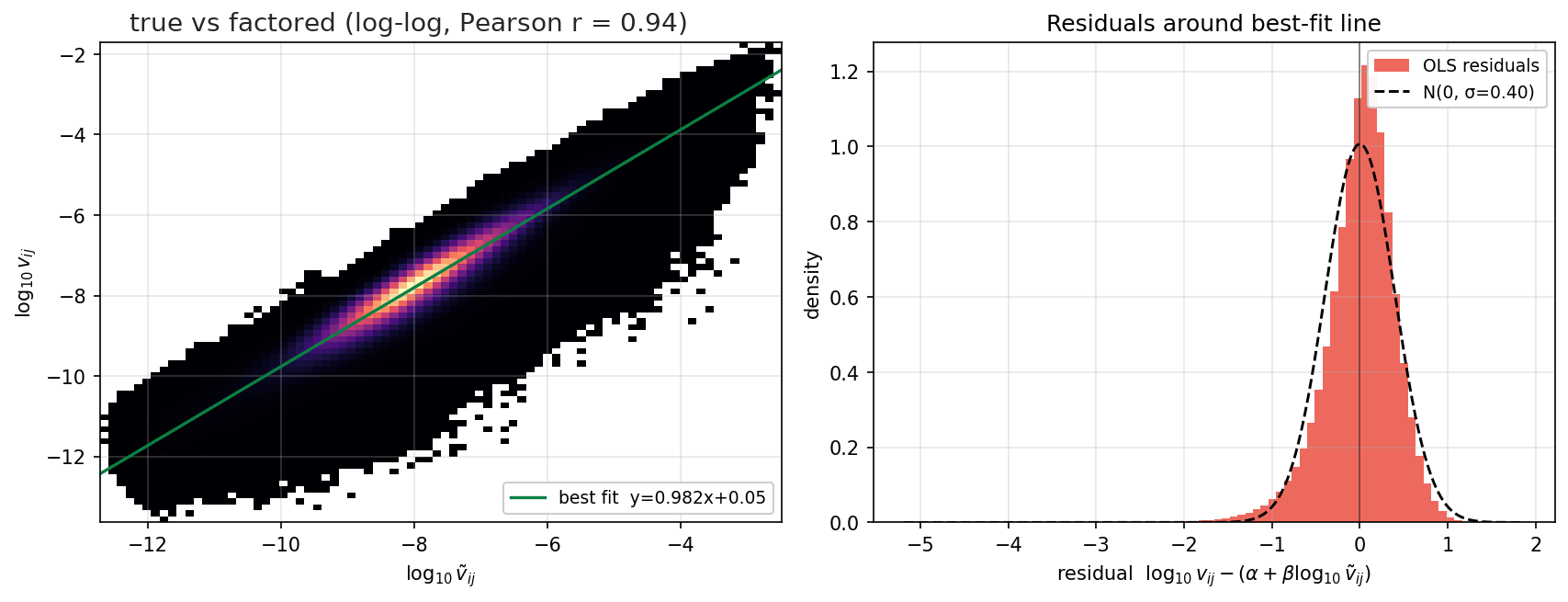}
 \caption{\textbf{A factored approximation to Adam's dense second moment on token tables.} %
 \textit{Left}: each point is one embedding parameter $(i,j)$; the factored $\tilde v_{ij}$ tracks Adam's dense $v_{ij}$. Pointwise, the match holds across ten orders of magnitude ($n = 38$M entries) with a correlation of $0.94$ and OLS slope $0.98$. %
 \textit{Right}: log-space residuals concentrate near zero with no heavy tail, illustrating the approximation error. Analysis in Section~\ref{app:decomposing_squared_gradient}.}
\label{fig:outerproduct_residual}
\end{figure}

Ember approximates the EMA of $G\odot G$ using its row and column means. We test this approximation directly: over all 38M embedding parameters of a 500-step GPT-2-small run on FineWeb, the factored estimate tracks Adam's dense second moment with correlation $0.94$ (Fig.~\ref{fig:outerproduct_residual}).

\section{Related work}

SGD \cite{rumelhart1986learning} is the simplest form of gradient descent, and uses the backprop gradient scaled by the learning rate to update parameters. 
Adam \cite{kingma2014adam} builds on it by dividing this by its RMS, and maintains a bias-corrected EMA of the 1st and 2nd moments which takes $\mathcal{O}(2VD)$ memory per token table.

Lion \cite{chen2023symbolic} shows momentum on just the gradient sign is competitive.
Muon \cite{jordan2024muon} forces singular values of the gradient matrix for each dense linear layer to be unitary, and is used by frontier models.
Shampoo \cite{gupta2018shampoo,anil2020scalable} maintains full row and column Gram matrices and preconditions with $L^{-1/4} G R^{-1/4}$.
Adafactor \cite{shazeer2018adafactor} factorizes Adam's second moment and removes first-moment state. The all-row variant of Ember coincides with their earlier Algorithm~2 under matched settings and away from numerical guards; active-row column averaging in our headline experiments generally changes the update. Their final recipe additionally uses relative steps, parameter scaling, update clipping, and scheduled decay. They also discuss token-frequency effects in embeddings (Sec.~3.2). Here we evaluate selective use on embedding and LM-head matrices in modern language-model training, with the body optimizer unchanged, and provide a distributed implementation. Our comparison with the final native recipe is in Appendix~\ref{sec:adafactor_comparison}.

SM3 \cite{anil2019memory} also uses row/column-sized buffers, but takes a minimum over their estimates rather than an outer product. Coupled Adam \cite{stollenwerk2025coupled} studies embedding-specific corrections to Adam's second moment.

Other memory-saving approaches include quantizing optimizer states (8-bit Adam \cite{dettmers2021eightbit}), low-rank gradient projection (GaLore \cite{zhao2024galore}), and sharing second-moment estimates within parameter blocks (Adam-mini \cite{zhang2024adammini}). We do not evaluate these methods here; our measured comparisons are against AdamW and the native Adafactor recipe.

\cite{lau2026symmetry} claims that linear layers are fundamentally symmetric, so a good optimizer should leverage this rather than update its parameters as if they were a 1D concatenated vector.
Modula \cite{bernstein2024modular, bernstein2025manifolds} was one of the first to present a grand unified theory of optimizers, mathematically grounded in norm duality. They derive that row-normalization is optimal for embeddings, corroborating the analysis of this paper.

\section{Discussion}

We provide a distributed version of Ember compatible with ZeRO \cite{rajbhandari2020zero}. Because Ember's optimizer state is $O(V+D)$, the embedding table optimizer state no longer requires sharding, simplifying distributed training setups.

\subsection{Practical implications}\label{app:practical}

\textbf{Faster iteration.}
The overwhelming majority of experimentation and finetuning is done on models $\leq 7$B
parameters where Adam's embedding state is a genuine binding constraint. For example, for Qwen2.5-7B, Adam stores roughly $8.72\,\mathrm{GB}$ VRAM for the token interface alone. Ember reduces this to about $1.2\,\mathrm{MB}$. 
Lower optimizer-state memory can make single-device experiments more practical. In a separate modded-nanoGPT experiment using Ember only on the value-embedding tables, peak allocated VRAM fell by approximately $2.9$\,GiB in each of three comparisons. This is a different setup from the input-embedding/LM-head state accounting above. In the FineWeb-Edu Pythia-410M runs, median tokens per second over the final 90\% of training differed by less than $0.5\%$ between Ember and AdamW in each of the three seed pairs.

\textbf{Cleaner engineering.} The optimizer state for LayerNorm and the embedding tables are all 1-D and can fit on a single GPU across even the largest models. Only linear layers/MoE require sharding, which can be done more simply as they scale cleanly with model depth and number of experts per layer.

\textbf{Multiple embedding tables.} The savings scale with $V \!\cdot\! D$, substantial for large-vocabulary or multimodal models where modalities each carry their own embedding table. Google's recent Gemma 3n models include Per-Layer Embedding \cite{gemma3n2025}, where each layer gets its own embedding table; presumably, these would have significant optimizer state and Ember would be even more useful there.

\subsection{Conclusion} 

Ember reduces token-interface optimizer state from $O(2VD)$ to $O(V+D)$. In our 50k-step pretraining experiments, it improves validation loss over tuned AdamW on both datasets. The fine-tuning and RL results provide additional evidence for using this low-state update on token tables while retaining the body optimizer.

\section*{Reproducibility Statement}
Experimental protocols and additional analyses are documented in the appendix. Algorithm~\ref{alg:ember} specifies the active-row column-averaging implementation used in the headline 50k-step experiments; the all-row mathematical variant is distinguished in Appendix~\ref{sec:adafactor_comparison}.

\section*{AI Use Statement}
Generative AI tools were used to help refine hypotheses, design and debug experiments, implement code, analyze and interpret results, survey related work, produce figures, and organize and edit the manuscript. The author reviewed the AI-assisted work and takes responsibility for the final content, including all text, claims, code, analyses, and artifacts.

\clearpage
\appendix
\section{Supplementary Material}

We begin with some preliminaries on information geometry:

\textbf{The Fisher metric.} Under regularity assumptions, the expected negative Hessian of the log likelihood equals the score second moment, with both expectations taken under the model:
\[
-\mathbb{E}_{p_\theta} \left[ \frac{\partial^2}{\partial \theta_i^2} \log p_\theta(x) \right] = \mathbb{E}_{p_\theta} \left[ \left( \frac{\partial}{\partial \theta_i} \log p_\theta(x) \right)^2 \right].
\]
This is an expectation identity, not an equality between derivatives for each example \cite{amari2016information,ly2017tutorial,nielsen2020elementary} (App.~\ref{app:fisher_curvature}).

\paragraph{Canonical Geometry of Cross-Entropy.}

The Fisher metric is especially elegant because it is not merely useful as an approximation, but it is the exact second-order term of the Kullback--Leibler divergence:
\[
D_{\mathrm{KL}}
\left(
p_\theta
\;\|\;
p_{\theta+d\theta}
\right)
=
\frac{1}{2}
d\theta^\top
F(\theta)
d\theta
+
\mathcal{O}(\|d\theta\|^3).
\]
This local geometry motivates natural-gradient methods \cite{amari1998natural,kakade2001natural,pascanu2014revisiting,desjardins2015natural}; it does not by itself establish which preconditioner works best in training.

\paragraph{Which second moment?}
The model Fisher averages score outer products under the model distribution. The empirical Fisher averages loss-gradient outer products over observed examples. Adam and Ember instead track an EMA of squared aggregated minibatch gradients. These objects need not coincide \cite{kunstner2019limitations}. We use Fisher geometry as motivation, not as an identity for Ember's buffer.

\subsection{When Fisher agrees with curvature}
\label{app:fisher_curvature}
Under standard regularity conditions, the model Fisher equals the expected negative Hessian of the log likelihood, with the expectation taken under the model. Relating this to the data-loss Hessian additionally requires agreement between the model and data distributions. Stationarity alone,
\begin{equation}
\mathbb{E}_{\mathrm{data}}[\nabla_\theta \ell]=0,
\label{eq:near_opt_score}
\end{equation}
does not establish that agreement. Our experiments do not require a Fisher--Hessian identity.

\subsection{Second-moment normalization}\label{app:sqrt_fisher}
Adam divides its first-moment estimate by the square root of a second-moment estimate. Ember uses the current gradient and a factored second-moment estimate instead. Both normalize gradient scale; neither operation by itself establishes a natural-gradient or trust-region interpretation.

\subsection{Optimizing sparsely activated parameters}\label{app:scaling_views}
For an input embedding, row $i$ receives contributions only when token $i$ appears. Under the coherent-mean regime described in Appendix~\ref{app:inverse_probability_curvature}, the expected row gradient scales with $p_i$ and its root second moment also scales with $p_i$. Dividing by the latter can therefore reduce frequency-dependent differences in update scale.

The three update scalings follow directly. If $g_i\propto p_i$ and its second-moment estimate $v_i\propto p_i^2$, then $g_i\propto p_i$, $g_i/v_i\propto 1/p_i$, and $g_i/\sqrt{v_i}\propto p_i^0$. The last ratio is dimensionless, but frequency cancellation alone does not establish isotropic gradient covariance or an identity with the model Fisher.

This argument is conditional: when the mean contribution is zero or the variance term dominates, the same cancellation does not follow. Ember estimates the observed second moment rather than imposing an inverse-frequency correction. The LM head receives dense softmax gradients and needs a separate analysis (App.~\ref{app:lmhead_smoothed}).

\subsection{Further work}
Future work could test which other parameter classes benefit from low-state second-moment approximations and under what training regimes.

\subsection{Experimental details}
\label{app:exp_details}

\textbf{Long-horizon Pythia-410M pretraining.}
\label{app:long_horizon_410m}
All runs use global batch size 32, sequence length 1024, and a 1,000-step warmup followed by cosine decay to 10\% of the peak learning rate. The Transformer body uses the same AdamW configuration in every run (learning rate $3\!\times\!10^{-4}$, $\beta=(0.9,0.95)$, and weight decay $0.1$), while the token tables use zero weight decay. For each corpus, AdamW and Ember receive the same token-interface learning-rate grid, $\{5\!\times\!10^{-4},10^{-3},2\!\times\!10^{-3},4\!\times\!10^{-3}\}$, in pilot runs before evaluation on three fresh paired seeds. The frozen endpoint is the mean of the 11 validation measurements at steps 45{,}000, 45{,}500, $\ldots$, 50{,}000. OpenWebText uses a tuning shard disjoint from its final training shards and an untouched validation shard.

\begin{table}[t]
\centering
\caption{Pythia-410M long-horizon pretraining. Losses are frozen endpoint means over three paired seeds; $\Delta$ is Ember minus AdamW in nats, with a paired 95\% $t$ interval.}
\label{tab:long_horizon_410m}
\small
\setlength{\tabcolsep}{4pt}
\begin{tabular}{lccccc}
\toprule
Corpus & AdamW LR & Ember LR & AdamW & Ember & $\Delta$ [95\% CI] \\
\midrule
FineWeb-Edu & $2\!\times\!10^{-3}$ & $10^{-3}$ & 3.09699 & \textbf{3.01677} & $\mathbf{-0.08021}$ $[-0.08646,-0.07396]$ \\
OpenWebText & $10^{-3}$ & $2\!\times\!10^{-3}$ & 3.21497 & \textbf{3.12680} & $\mathbf{-0.08817}$ $[-0.10238,-0.07395]$ \\
\bottomrule
\end{tabular}
\end{table}

At the shared token-table shape, Ember's two FP32 row-and-column buffers replace 205,942,528 FP32 AdamW state elements, removing 823,770,112 bytes (approximately 824 MB) of token-interface optimizer state.

These 50k-step runs use active-row column averaging and the numerical guards in Algorithm~\ref{alg:ember}. Earlier experiments use separate implementations; we do not assume that every reported experiment shares this averaging convention.

\textbf{Scaling parameter count.} All model-scaling runs finetune on Alpaca SFT for 250 steps with effective batch size 128, learning rate $2\times10^{-5}$ with cosine decay, and bf16 precision. The transformer body uses AdamW throughout; only the \emph{token-embedding / LM-head} optimizer is varied.

\begin{figure}[t]
\centering
\includegraphics[width=0.72\linewidth]{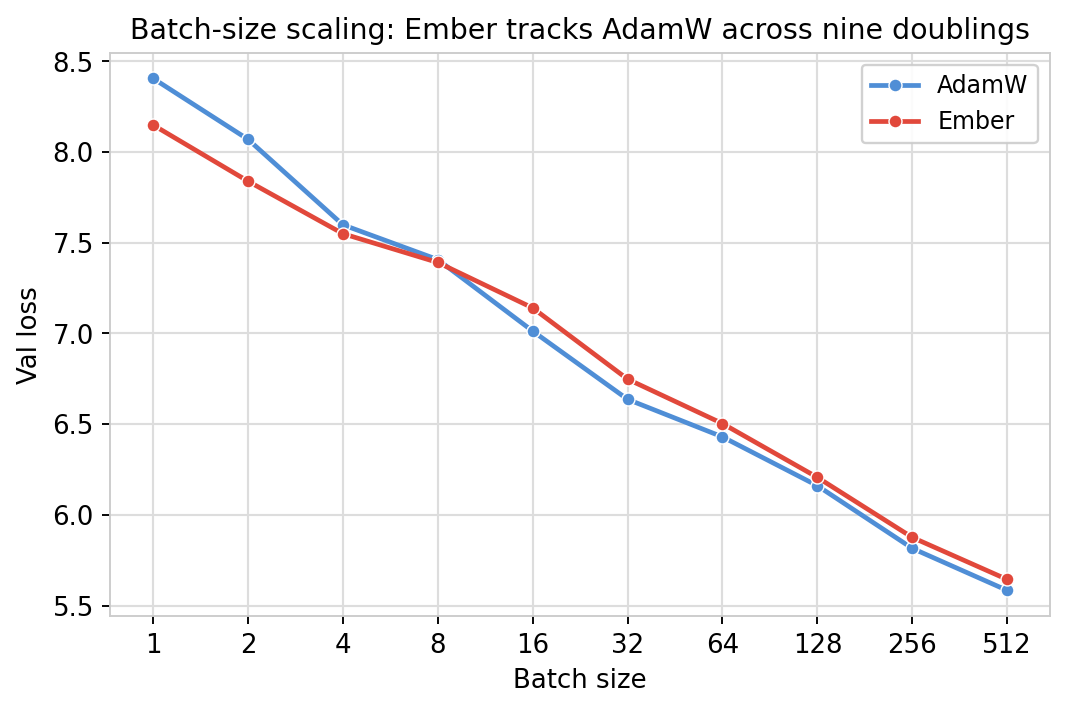}
        \caption{Batch-size scaling across nine doublings. Ember leads at the smallest batches, while AdamW leads at batches 16--256 ($n{=}3$ paired seeds; plotted curve is a representative run).}
        \label{fig:bs_scaling}
\end{figure}

\textbf{Scaling batch size.} We train GPT-2-small ($V=50257$, block 144) on FineWeb with both token tables untied, sweeping the batch size across nine doublings from $1$ to $512$ for 300 steps (3 paired seeds). The replicated comparison uses step 195, the same checkpoint as the earlier representative curve in Figure~\ref{fig:bs_scaling}; this measures early-training performance, not final convergence.

We find that Ember obeys the same near-linear law in log-batch across nine doublings
  (Figure~\ref{fig:bs_scaling}). Over $n{=}3$ paired seeds, Ember performs better at the smallest batches (paired $\Delta = -0.294 \pm 0.040$ at batch 1, $-0.159 \pm 0.077$ at batch 2), while AdamW has lower loss at batches 16--256.

\textbf{Autoregressive image generation.} Ember is competitive in computer vision tasks: on LlamaGen image-AR, using $1500\times$ less optimizer state, Ember is virtually indistinguishable from Adam (Fig.~\ref{fig:llamagen_panel}). Setup in Appendix~\ref{app:exp_details}.

LlamaGen is a vanilla decoder-only transformer that generates images autoregressively. Instead of predicting over discrete language tokens, it trains its own learnable embedding table to look up within a frozen VQGAN codebook. We compare Adam and Ember at the same fixed learning rate of $1\times10^{-4}$, rather than comparing independently tuned optima.

\subsection{Comparison to Adafactor}
\label{sec:adafactor_comparison}
The native Adafactor baseline uses the final recipe: relative steps, parameter scaling, RMS update clipping, and scheduled decay. Ember uses a fixed second-moment decay with explicit bias correction and an externally chosen learning rate. Table~\ref{tab:decomp} compares these complete recipes over 200 steps; it does not isolate which difference causes the loss gap. The 50k-step experiments compare Ember with AdamW, not Adafactor.

\textbf{Relation of the all-row variant to Algorithm 2.}
For the variant that averages columns over all $V$ rows, let $q_t=1-\beta_2^t$ and let $m_t$ be the common grand mean of the uncorrected row and column EMAs. Algorithm~2 in \cite{shazeer2018adafactor} stores sums rather than means, so its buffers satisfy $R_t=D r_t$ and $C_t=V c_t$. Away from numerical guards, both reconstructions give
\[
\tilde v_{t,ij}
=\frac{r_{t,i}c_{t,j}}{q_t m_t}.
\]
With the same gradient history, zero initialization, decay, learning rate, and epsilon convention, the all-row parameter updates agree. This identity assumes $m_t>0$ and inactive numerical floors. It does not generally hold for the active-row column average used in our 50k-step runs. Algorithm~\ref{alg:ember} guards an all-zero history explicitly.

\begin{table}[h]
\centering
\caption{Recipe comparison and component ablations: validation loss at step 200 (GPT-2-small / FineWeb, Muon on
linear layers), mean $\pm$ across-seed sample sd over $3$ paired seeds; $\Delta$ is the paired
per-seed difference versus AdamW. Canonical Adafactor uses its native recipe (relative step
size, update clipping, $\beta_2$ schedule).}
\label{tab:decomp}
\small\setlength{\tabcolsep}{4pt}
\begin{tabular}{lcc|cc}
\toprule
 & \multicolumn{2}{c|}{batch 8} & \multicolumn{2}{c}{batch 256} \\
Token-table optimizer & val@200 & $\Delta$ & val@200 & $\Delta$ \\
\midrule
AdamW & 7.354{\scriptsize$\pm$.030} & --- & 5.829{\scriptsize$\pm$.002} & --- \\
Canonical Adafactor & 7.302{\scriptsize$\pm$.030} & $-.052{\scriptstyle\pm}.032$ & 6.093{\scriptsize$\pm$.024} & $+.264{\scriptstyle\pm}.026$ \\
Ember & 7.207{\scriptsize$\pm$.023} & $-.147{\scriptstyle\pm}.032$ & 5.842{\scriptsize$\pm$.020} & $+.013{\scriptstyle\pm}.021$ \\
Ember + first-moment EMA & 7.237{\scriptsize$\pm$.031} & $-.117{\scriptstyle\pm}.040$ & 5.813{\scriptsize$\pm$.030} & $-.016{\scriptstyle\pm}.033$ \\
Ember row-only & 7.312{\scriptsize$\pm$.021} & $-.042{\scriptstyle\pm}.030$ & 5.890{\scriptsize$\pm$.017} & $+.061{\scriptstyle\pm}.016$ \\
\bottomrule
\end{tabular}
\end{table}

Ember improves over the tested native recipe by $0.095$ nats at batch 8 and $0.251$ nats at batch 256. The separate bias-correction ablation measures its effect within Ember.

\subsection{Supplementary figures}

The following figures compare the measured principal-component projections with low-degree polynomial fits.

\begin{figure}[H]
\centering
\includegraphics[width=\linewidth]{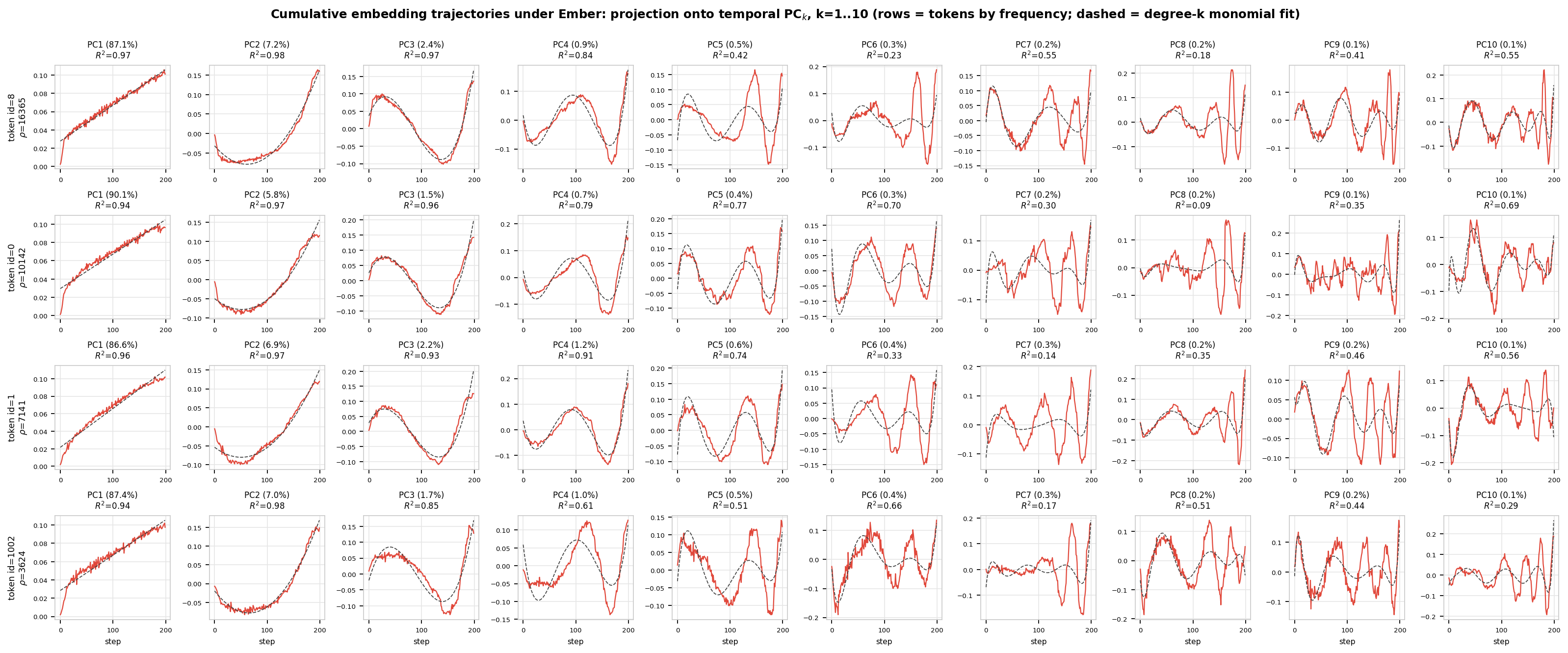}
\caption{For each of four monitored tokens, we stack its embeddings over training and compute the SVD. Each panel shows a principal-component projection. The observed projections resemble low-degree polynomials, and the leading component explains roughly $90\%$ of trajectory variance.}

\label{fig:trajectory_pca}
\end{figure}

\begin{figure}[H]
\centering
\includegraphics[width=0.72\linewidth]{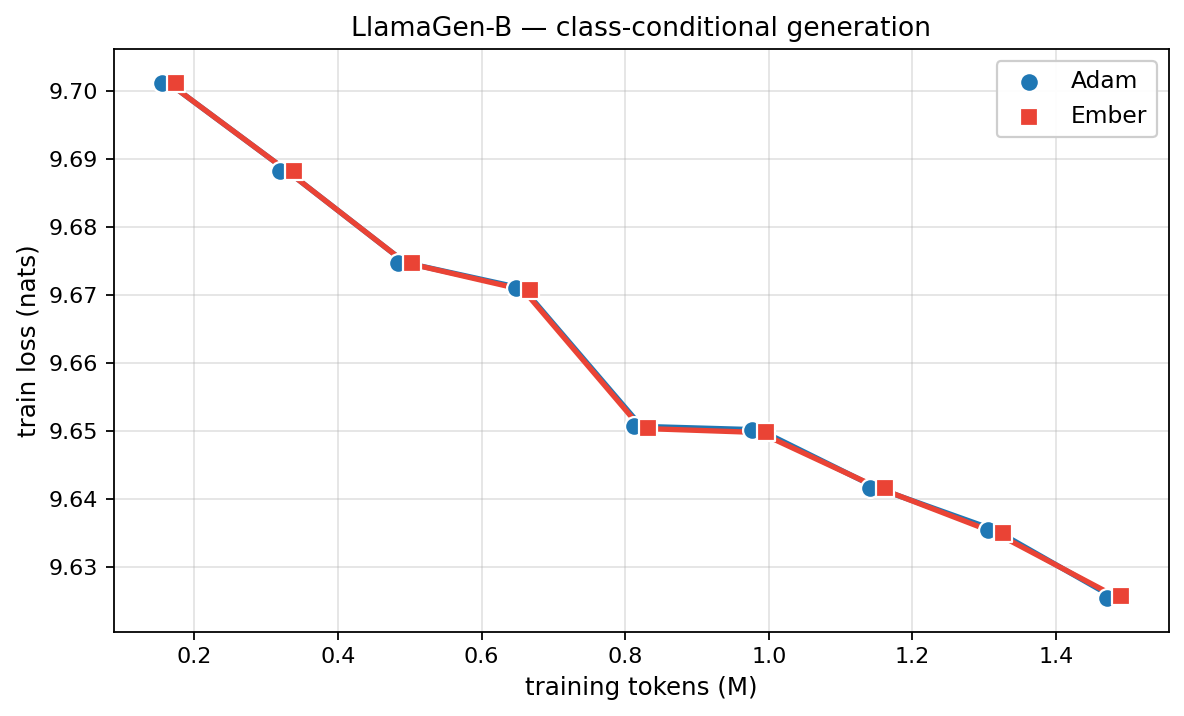}
        \caption{Autoregressive image generation results. Ember matches Adam while using substantially less optimizer state.}
        \label{fig:llamagen_panel}
\end{figure}

\subsection{Ablations}

\textbf{Nesterov look-ahead ablation.}
We ran a $3 \times 3$ grid over $\beta_2 \in \{0.999, 0.95, 0.9\}$ and learning rate $\mathrm{lr} \in \{5\times10^{-4}, 10^{-3}, 2\times10^{-3}\}$, and consistently found that injecting the current $g^2$ estimate into the denominator hurts performance across all learning rates. Intuitively, the second moment acts as a smooth scale/curvature estimate and should not react to instantaneous gradients. Canonical $\beta_2 = 0.999$ was already optimal.

\subsection{Token frequency and squared batch gradients}
\label{app:inverse_probability_curvature}

Consider an input-embedding row $i$. For each of $M$ independent positions, let $I_a$ indicate token $i$ and $u_a$ denote its gradient contribution when selected. Write
\[
\Pr(I_a=1)=p_i,\qquad
\mu_i=\mathbb{E}[u_a\mid I_a=1],\qquad
S_i=\mathbb{E}[u_a u_a^\top\mid I_a=1].
\]
For the mean-reduced batch gradient $\bar g_i=M^{-1}\sum_a I_a u_a$,
\[
\mathbb{E}[\bar g_i]=p_i\mu_i,\qquad
\mathbb{E}[\bar g_i\bar g_i^\top]
=\frac{p_i}{M}S_i+\frac{M-1}{M}p_i^2\mu_i\mu_i^\top.
\]
The first term is linear in frequency. The second is quadratic and requires a nonzero conditional mean. Equivalently, writing $S_i=\operatorname{Cov}(u_a\mid I_a=1)+\mu_i\mu_i^\top$, the count-variance term is $p_i(1-p_i)\mu_i\mu_i^\top/M$. For batch-summed gradients, multiply the second-moment expression by $M^2$.

The quadratic term can dominate for frequent tokens with sufficiently aligned contributions; it vanishes when $\mu_i=0$. Independence across positions is an approximation for language batches. Our measured association between row-gradient energy and frequency ($r=0.976$) motivates row-wise scaling, but does not identify an exponent. Ember measures the squared gradients directly and does not assume a global $p_i^2$ law.

\subsection{Outer Product Unit Alignment}
\label{app:outer_product_unit_alignment}

For the all-row variant, let $A_t$ be the bias-corrected EMA of $G_t\odot G_t$. Its factors are the row means $\hat r$ and column means $\hat c$ of $A_t$. Both grand means equal $\bar A_t$:
\[
\operatorname{mean}_i\hat r_i
=\operatorname{mean}_j\hat c_j
=\bar A_t.
\]
Away from numerical floors, the geometric-mean normalizer gives
\[
s_t=\sqrt{\operatorname{mean}(\hat r)\operatorname{mean}(\hat c)}
=\bar A_t,\qquad
\tilde v_{ij}=\frac{\hat r_i\hat c_j}{\bar A_t}.
\]
Each factor has units $[g]^2$, so division by $\bar A_t$ restores the units of a second moment.

If $A_{t,ij}=a_i b_j$ is exactly rank one with positive grand mean, then $\hat r_i=a_i\bar b$, $\hat c_j=\bar a b_j$, and $\bar A_t=\bar a\bar b$, giving $\tilde v_{ij}=A_{t,ij}$.

These equalities apply to all-row averaging. With active-row column averaging, the two grand means generally differ, and exact rank-one reconstruction is not guaranteed. The geometric mean still restores second-moment units away from its numerical floor, but this dimensional argument does not establish equivalence to Adafactor or exact reconstruction.

\begin{figure}[H]
\centering
\includegraphics[width=\linewidth]{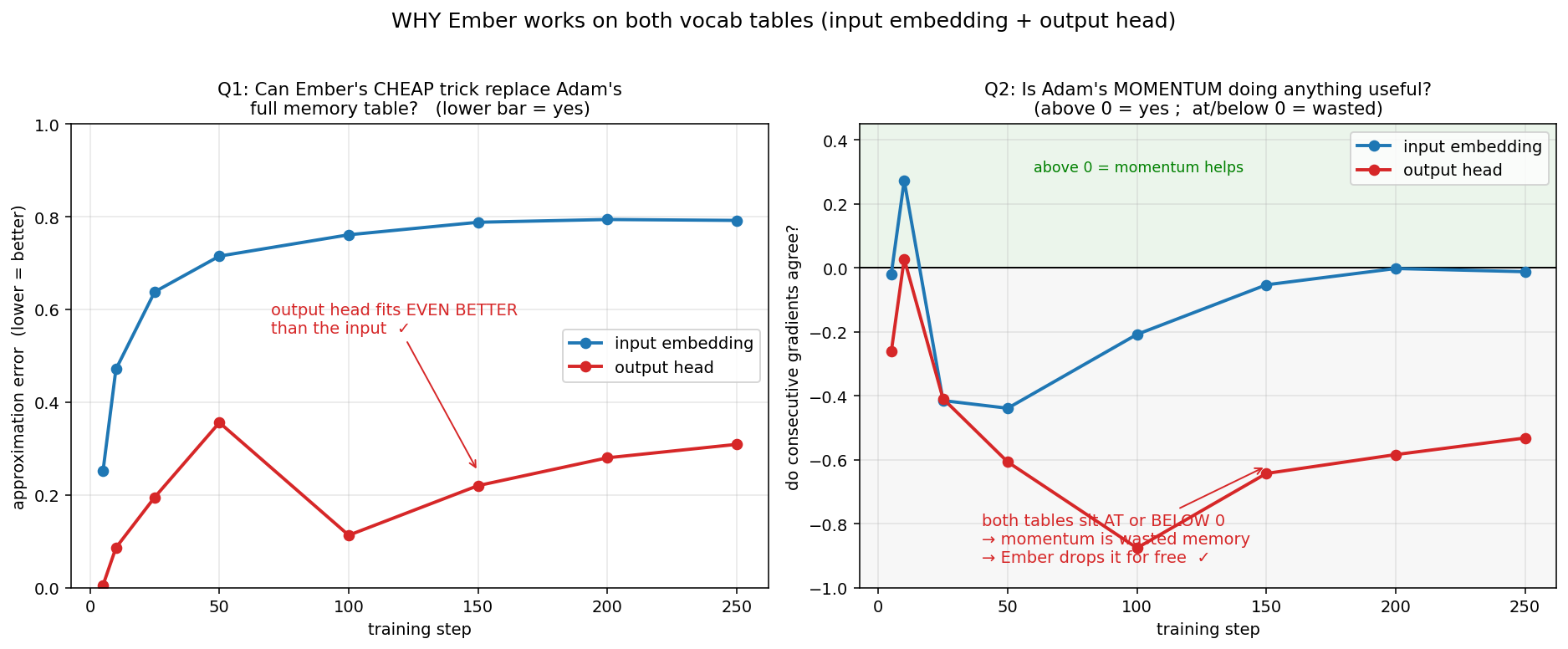}
\caption{\textbf{Gradient diagnostics on token tables.}
We instrument gradients of the untied input embedding $E_{\mathrm{in}}\in\mathbb{R}^{V\times d}$ and output head $W_{\mathrm{out}}\in\mathbb{R}^{V\times d}$ during GPT-2 small training on FineWeb.
Left: Ember's row-column factored second moment approximates the dense squared-gradient structure, where the y-axis is relative error.
Right: temporal gradient autocorrelation is near zero or negative. Table~\ref{tab:decomp} directly evaluates the effect of adding first-moment momentum.}
\label{fig:no_first_mom}
\end{figure}

\subsection{Related normalization methods}
\label{app:muon_polar}

\subsubsection{Muon: spectral normalization}
For a matrix $H$ with compact SVD $H=U_r\Sigma_r V_r^\top$ over its nonzero singular values,
\[
H(H^\top H)^{\dagger/2}=U_rV_r^\top,
\]
where $\dagger/2$ denotes the inverse square root on the nonzero spectrum. This is the polar normalization used in Muon, typically applied to a momentum-adjusted gradient and approximated numerically. It normalizes singular values rather than individual coordinates. The Gram matrix $H^\top H$ is not the full Fisher block on the vectorized parameters.

\subsubsection{Shampoo: two-sided preconditioning}
Shampoo \cite{gupta2018shampoo,anil2020scalable} accumulates row and column Gram matrices,
\[
L=\sum_t G_tG_t^\top,\qquad R=\sum_t G_t^\top G_t,
\]
and preconditions by $L^{-1/4}GR^{-1/4}$, with regularization for singular matrices. Vectorization gives $(R\otimes L)^{-1/4}\operatorname{vec}(G)$ for symmetric positive-definite factors. This is a structured preconditioner, not in general the same Fisher conditioner as Adam or Ember.

\subsection{LM-head gradients}
\label{app:lmhead_smoothed}

For hidden state $h_a$, target $y_a$, and predicted token probability $q_{a,i}$, the batch-summed gradient of LM-head row $i$ is
\[
G_i=\sum_{a=1}^{M}\left(q_{a,i}-\mathbf{1}[y_a=i]\right)h_a.
\]
Unlike input embeddings, each row receives a softmax contribution even when its token is not the target. The target and softmax terms can cancel, so a target-count argument alone does not determine the full gradient's frequency scaling.

Under independent positions, let $z_{a,i}=(q_{a,i}-\mathbf{1}[y_a=i])h_a$ have mean $\nu_i$ and covariance $\Sigma_i$. Then
\[
\mathbb{E}[G_iG_i^\top]=M\Sigma_i+M^2\nu_i\nu_i^\top.
\]
This separates variance and coherent-mean contributions, but does not imply rank-one structure across vocabulary and feature dimensions. Using row-column moments for the LM head is an empirical approximation evaluated in our untied-token-table experiments.

  \begin{figure}[t]
    \centering
    \caption{Earlier embedding-optimizer ablations at two batch sizes for GPT-2 small / FineWeb. The batch-8 row-only run uses learning rate $2\times10^{-3}$, versus $10^{-3}$ in the three-seed comparison in Table~\ref{tab:decomp}; the two row-only values are not estimates under the same settings. The row-only version is close to Ember at batch 256 but worse at batch 8; column-only is worse at both batches.}
    \label{fig:ember_ablations}
    \begin{tabular}{lccc}
    \toprule
    Embedding optimizer & Val@200 (bs=8) & Val@200 (bs=256) & State \\
    \midrule
    AdamW & 7.325 & 5.897 & $2VD$ \\
    Muon & 7.354 & 6.250 & $VD$ \\
    Adafactor & 7.311 & 6.137 & $V{+}D$ \\
    \textbf{Ember} & \textbf{7.219} & \textbf{5.890} & $V{+}D$ \\
    \midrule
    Ember row-only & 7.665 & 5.911 & $V$ \\
    Ember col-only & 10.899 & 7.958 & $D$ \\
    \bottomrule
    \end{tabular}
  \end{figure}

\begin{figure}[!htbp]
\centering
\includegraphics[width=\linewidth]{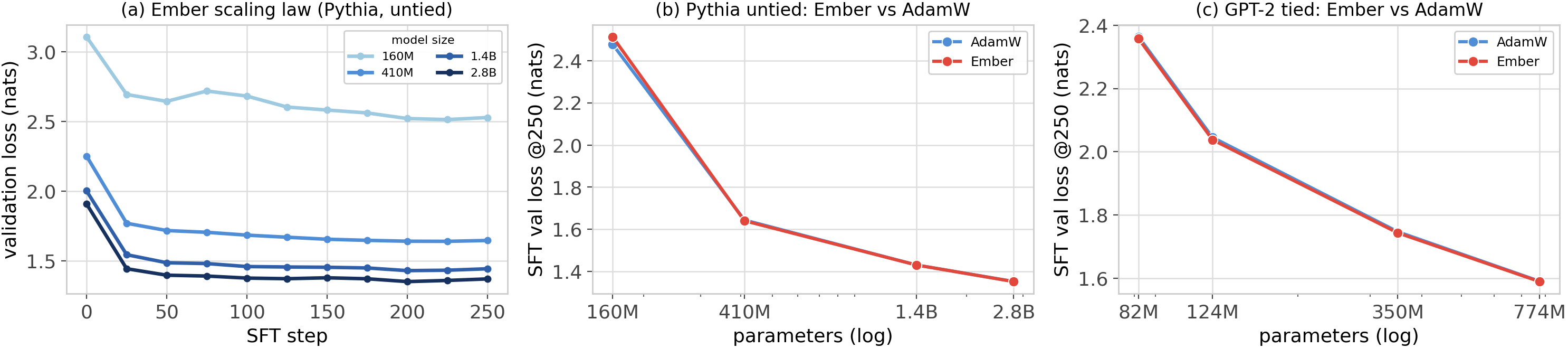}
\caption{\textbf{Model-scaling results for Ember on token tables.} Validation loss after 250 supervised fine-tuning steps. $n{=}3$ paired seeds per point. Training setup in Appendix~\ref{app:exp_details}.}
\label{fig:model_scaling}
\end{figure}
\subsection{Distributed Ember}

\textbf{Replicate instead of shard.} Adam's $\mathcal{O}(2VD)$ fp32 state is roughly $4\times$ larger than the bf16 table it serves, so frameworks like ZeRO/FSDP split it across devices and every checkpoint records that split, making resumption on a different cluster size a resharding exercise. Ember's $\mathcal{O}(V{+}D)$ state is a few hundred KB: each device keeps a full copy, and a checkpoint is one small file that loads anywhere.

\textbf{The table can skip sharding entirely.} Since the optimizer state is what usually forces token tables into sharded training, removing it lets the table's weights and gradients train fully replicated at moderate scale.

\textbf{Persistent state and working memory.} Only the $V{+}D$ factors persist between steps. Materializing the denominator still needs $O(VD)$ temporary memory, so the state bound is not a bound on peak VRAM. Input-embedding gradients can be row-sparse; LM-head gradients are generally dense. The algorithm does not require sparse execution.

\subsection{Optimization trajectories}\label{app:opt_landscape}

\begin{figure}[H]
\centering
\includegraphics[width=0.82\linewidth]{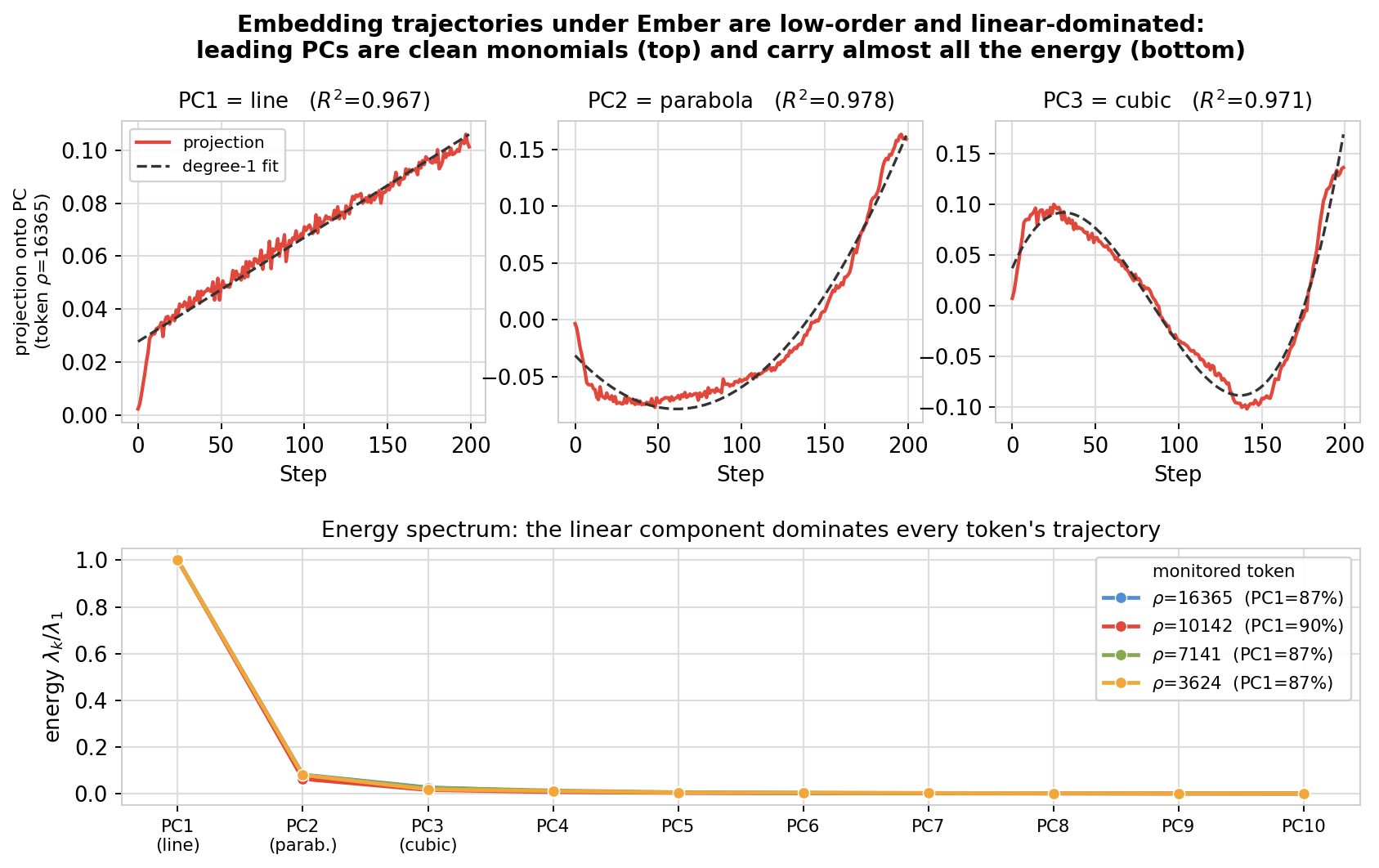}
\caption{\textbf{Polynomial fits to token optimization trajectories.} Fits to the first three principal-component projections and their energy spectrum. The leading component dominates in these runs. More examples are in Figure~\ref{fig:trajectory_pca}.}

\label{fig:pca_first_3}
\end{figure}

We stack each monitored token's embedding over training into a $D\times T$ matrix and examine its singular vectors (Fig.~\ref{fig:pca_first_3},~\ref{fig:trajectory_pca}). In these runs, the leading component explains roughly $90\%$ of trajectory variance, and several component projections resemble low-degree polynomials. This describes the measured parameter paths; it does not establish that the loss landscape is one-dimensional or free of local optima.

